\pdfoutput=1
\documentclass[10pt,twocolumn,letterpaper]{article}

\usepackage{cvpr}              % To produce the CAMERA-READY version
\usepackage{graphicx}
\usepackage{amsmath}
\usepackage{amssymb}
\usepackage{booktabs}
\usepackage[accsupp]{axessibility}
\usepackage[belowskip=-10pt,aboveskip=5pt]{caption}

\usepackage[pagebackref,breaklinks,colorlinks]{hyperref}
\usepackage[dvipsnames]{xcolor}

\usepackage[capitalize]{cleveref}
\crefname{section}{Sec.}{Secs.}
\Crefname{section}{Section}{Sections}
\Crefname{table}{Table}{Tables}
\crefname{table}{Tab.}{Tabs.}

\def\cvprPaperID{*****} % *** Enter the CVPR Paper ID here
\def\confName{CVPR}
\def\confYear{2023}

\begin{document}


\title{Agronav: Autonomous Navigation Framework for Agricultural Robots and Vehicles using Semantic Segmentation and Semantic Line Detection}

% \author{Shivam Kumar Panda\\
% {\tt\small shivamkp@g.ucla.edu}
% \and
% Yongkyu Lee\\
% {\tt\small yongkyulee@g.ucla.edu}
% \and
% M. Khalid Jawed\\
% {\tt\small khalidjm@seas.ucla.edu}
% }
\author{Shivam K Panda* \hspace{1.5cm} Yongkyu Lee* \hspace{1.5cm} M. Khalid Jawed\\
Dept. of Mechanical and Aerospace Engineering, University of California Los Angeles\\
{\tt\small \{shivamkp, yongkyulee\}@g.ucla.edu, khalidjm@seas.ucla.edu  }
}
\maketitle
\def\thefootnote{*}\footnotetext{These authors contributed equally to this work}

%%%%%%%%% ABSTRACT
\begin{abstract}
The successful implementation of vision-based navigation in agricultural fields hinges upon two critical components: 1) the accurate identification of key components within the scene, and 2) the identification of lanes through the detection of boundary lines that separate the crops from the traversable ground. We propose Agronav, an end-to-end vision-based autonomous navigation framework, which outputs the centerline from the input image by sequentially processing  it through semantic segmentation and semantic line detection models. We also present Agroscapes, a pixel-level annotated dataset collected across six different crops, captured from varying heights and angles. This ensures that the framework trained on Agroscapes is generalizable across both ground and aerial robotic platforms. Codes, models and dataset will be released at \href{ https://github.com/StructuresComp/agronav/}{github.com/StructuresComp/agronav/}.
\end{abstract}

%%%%%%%%% BODY TEXT
\section{Introduction}
\label{sec:intro}

According to World Food Programme, the population of those experiencing food insecurity is projected to be 342.5 million in 2023, which is more than double the same population in 2020. This concerning trend can be attributed to population growth~\cite{brown2019}, climate change~\cite{verschuur2021}, labor shortage~\cite{laborde2020}, and food affordability driven by high fertilizer prices~\cite{bjornlund2022}. To address these challenges, the concept of Precision Agriculture has emerged as a sustainable solution, leveraging modern technology to optimize crop and livestock management practices~\cite{CISTERNAS2020, monteiro2011, shafi2019}. A key aspect of Precision Agriculture is automation technology, which aims to minimize resource expenditure while maximizing efficiency. Central to automation technology is autonomous navigation, which enables robotic platforms to operate in the field without human intervention.

Some existing methods of achieving autonomous navigation in agricultural fields rely on real-time kinematic GNSS (RTK-GNSS)~\cite{thuilot2002automatic, bawden2017robot}, LiDAR~\cite{iqbal2020} and depth cameras~\cite{aghi2020}. Some limitations of RTK-GNSS equipment include its high cost, vulnerability to region-specific outages, reliance on geo-referenced auto-seeding, and signal attenuation problems for smaller mobile robots designed for under-canopy tasks~\cite{ahmadi2020visual}. In addition, the trajectories planned using waypoints with the RTK-GNSS system does not account for the dynamic, changing environment of the agricultural field~\cite{emmi2021}. This necessitates the use of onboard sensors to observe the changes in the environment in closer proximity. While LiDARs are an integral part of autonomous driving in urban environments, where most obstacles are defined by hard and simple surfaces, their use in the agricultural field is limited due to the natural differences in the environment. The complex and soft nature of the surrounding obstacles, such as leaves and stems and their close proximity to the UGV make it a hostile environment for using LiDARs. In addition, LiDARs entail complex tasks such as point cloud classification and multimodal fusion to extract meaning from the acquired data~\cite{lohar2021}.

\begin{figure}[t]
  \centering
    \includegraphics[width=\linewidth]{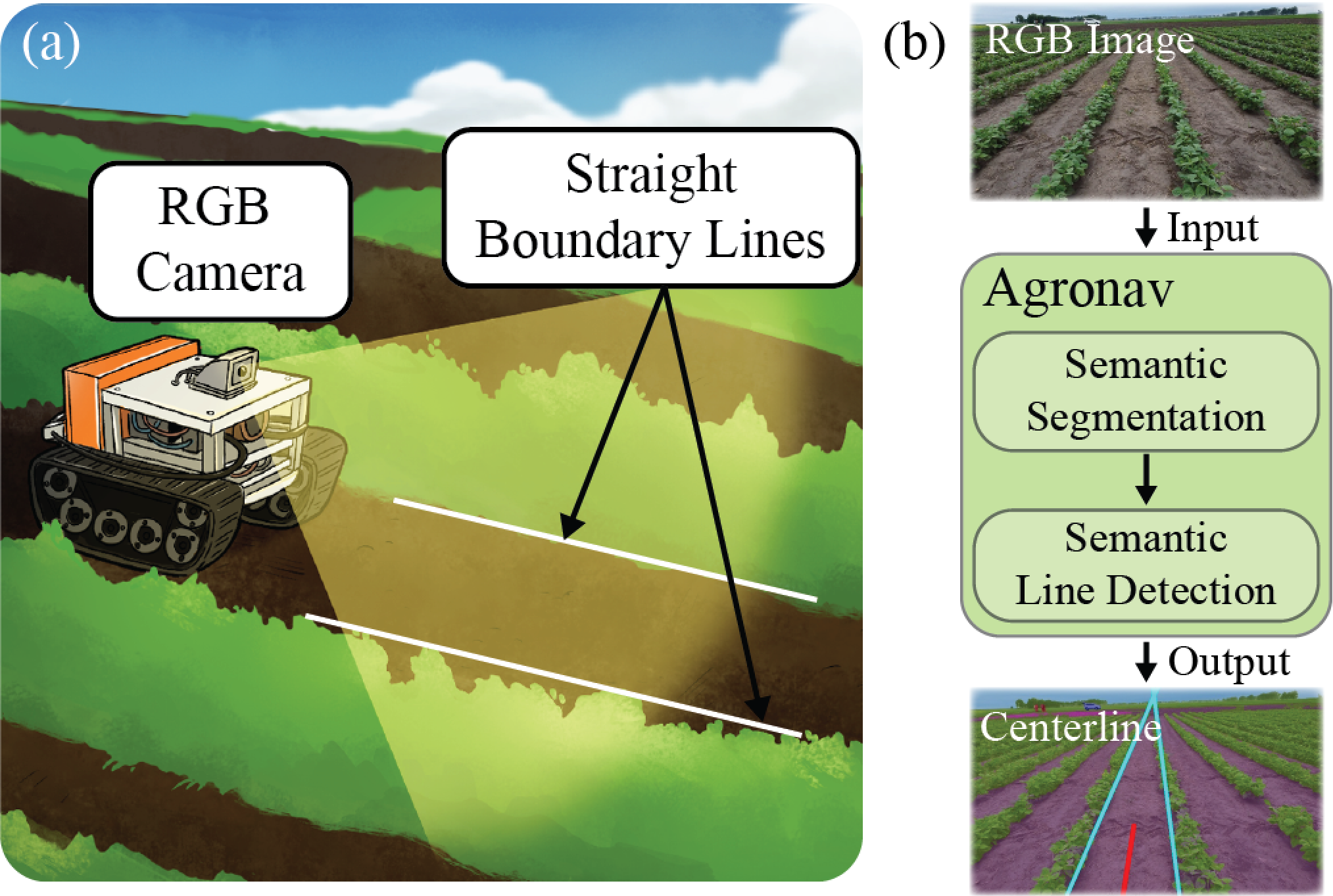}
   \caption{Overview of Agronav. (a) In an agricultural field, lanes are represented as straight boundary lines. A vision-based navigation framework can effectively capture these lines. (b) High-level pipeline of Agronav. Centerlines are extracted from input RGB images through a series of operations. }
   \label{fig:intro}
\end{figure}

To this end, we propose an end-to-end pipeline for autonomous navigation in the agriculture field centered around efficiency and simplicity. Our method avoids the use of expensive equipment such as RTK-GNSS and LiDAR and is entirely vision-based, which requires nothing but a single RGB camera (see Figure \ref{fig:intro}a). We frame the problem as a series of two downstream tasks: 1) semantic segmentation, which labels each pixel of the input RGB image as one of the predefined classes; and 2) semantic line detection, which extracts the two boundary lines from the overlayed image, an equally weighted blend of the raw image and the color mask, which is obtained as a result of the first task (see Figure \ref{fig:intro}b). The main contributions of our paper are as follows: 
\begin{itemize}
    \item We propose a simple and efficient end-to-end pipeline that extracts the centerline from an RGB image from a series of two operations: semantic segmentation and semantic line detection.
    \item We utilize domain adaptation with minimal annotated data for the semantic segmentation model by reorganizing the labels of a publicly available dataset, Cityscapes, and using accurate inferences from a large vision model, ViT-Adapter.
    \item We demonstrate that semantic line detection, which capitalizes on the structured environment of an agriculture field where crops are planted in straight rows, can successfully extract two boundary lines for various test cases.
    % \item We provide a study on different architectures for semantic segmentation, in order to find a good balance between performance and required computational resources, which is an important consideration for mobile platforms.
    \item We provide an open-source dataset, Agroscapes, which can be used as a benchmark for scene understanding in agricultural fields for different crops.
\end{itemize}

The rest of the paper is organized as follows. Section \ref{sec:relatedwork} provides a detailed review of semantic segmentation, semantic line detection, and autonomous navigation in agriculture, which are core related topics of our work. Section \ref{sec:dataset} introduces the data collection and annotation for the training of the semantic segmentation model and the line detection model. Section \ref{sec:methodology} covers the detailed methodologies related to our autonomous navigation pipeline. Section \ref{sec:result} provides the result that quantifies the performance of our navigation pipeline. Lastly, Section \ref{sec:conclusion} discusses the future direction of this research. 
%-------------------------------------------------------------------------

% Update the cvpr.cls to do the following automatically.
% For this citation style, keep multiple citations in numerical (not
% chronological) order, so prefer \cite{Alpher03,Alpher02,Authors14} to
% \cite{Alpher02,Alpher03,Authors14}.

% \begin{figure*}
%   \centering
%   \begin{subfigure}{0.68\linewidth}
%     \fbox{\rule{0pt}{2in} \rule{.9\linewidth}{0pt}}
%     \caption{An example of a subfigure.}
%     \label{fig:short-a}
%   \end{subfigure}
%   \hfill
%   \begin{subfigure}{0.28\linewidth}
%     \fbox{\rule{0pt}{2in} \rule{.9\linewidth}{0pt}}
%     \caption{Another example of a subfigure.}
%     \label{fig:short-b}
%   \end{subfigure}
%   \caption{Example of a short caption, which should be centered.}
%   \label{fig:short}
% \end{figure*}

%------------------------------------------------------------------------
\section{Related Works}
\label{sec:relatedwork}

\subsection{Semantic Segmentation in Autonomous Driving}\label{sec:semseg-ad}

Semantic segmentation is the process of assigning each pixel in an image to a particular object class. In autonomous driving, semantic segmentation is used to identify objects in the environment, such as roads, pedestrians, vehicles, traffic signs, etc. and to generate a detailed map of the scene. This information is then used to plan the vehicle's trajectory and ensure safe and efficient driving.

Over the years, several state-of-the-art models have been developed for semantic segmentation in autonomous driving. These models leverage deep learning techniques to achieve high accuracy and robustness in a variety of driving scenarios. One of the most popular models is the Fully Convolutional Network (FCN) proposed by Long \textit{et al.} \cite{long2015fully}, which replaces the fully connected layers of a traditional Convolutional Neural Network (CNN) with convolutional layers to enable pixel-wise predictions. Other popular models such as the U-Net model \cite{ronneberger2015u} uses a U-shaped architecture to capture both local and global features and has been shown to achieve high accuracy on medical image segmentation tasks. Deeplab model \cite{chen2017deeplab} on the other hand, uses atrous convolution to increase the receptive field of the network and improve segmentation accuracy.

These models have been trained end-to-end on annotated datasets such as the Cityscapes \cite{cordts2016cityscapes}, which contains high-resolution images of urban environments with pixel-level annotations for 30 object classes. Other 3D dataset have also been used in the autonomous driving community such as the KITTI dataset \cite{geiger2012we}, which contains images as well as point cloud data obtained from a moving vehicle.
% and the ApolloScape dataset \cite{huang2018apolloscape}, which contains high-resolution images and point cloud data of urban environments.

In recent years, transformer-based models such as Vision Transformer (ViT) \cite{dosovitskiy2020image} and Swin Transformer \cite{liu2021Swin} have shown remarkable performance on a variety of computer vision tasks including semantic segmentation. ViT utilizes the self-attention mechanism to capture global dependencies and is designed to work well on large-scale datasets. Swin Transformer introduces hierarchical structures with shifted windows to further improve the performance. Some other state-of-the-art models in this context are ViT-Adapter \cite{chen2022vision}, HRNet \cite{SunXLW19}, SegFormer \cite{xie2021segformer}, and ResNeSt\cite{zhang2020resnest}. 

\subsection{Semantic Line Detection}
Semantic lines are characteristic straight lines that capture the essence of a scene. Identifying these lines, which are often implied than obvious, can enhance understanding of an image. Besides the most prominent application in photographic composition to improve aesthetics, semantic lines are also critical in autonomous driving. In autonomous driving, the boundaries of road lanes and key road features serve as important semantic lines. Though many road features are represented as curved lines in urban settings, a typical agricultural field is more structured, where crops are planted in straight rows. This organized structure of the agricultural field simplifies the task of detecting lanes. Lanes can be approximated with straight lines, which are the boundaries that divide the crops from the traversable ground. 

The practice of detecting straight lines from images dates back to an image processing technique called the Hough Transform. More recently, success of CNNs in computer vision led to numerous deep learning-based approaches. A CNN-based semantic line detector named SLNet and open-source dataset SEL was proposed in ~\cite{Lee2017}. This work treated the identification of semantic lines as a combination of classification and regression tasks. An improvement of this study, which used the attention mechanism as well as matching and ranking, was introduced by Jin \textit{et al.}~\cite{jin2020}. The proposed line detection algorithm DRM consists of three neural networks: D-Net, R-Net and M-net. While the D-Net extracts semantic lines through the mirror attention module, R-Net, and M-net are Siamese networks that select the most meaningful lines and remove redundant lines. Most recently, Zhao \textit{et al.} introduced the Deep Hough Transform method, which is an end-to-end framework that combines convolutional layers for feature extraction and the Hough Transform to detect semantic lines~\cite{zhao2021}.

\subsection{Autonomous Navigation in Agriculture}

% One potential and easy solution to achieve navigation is to use Global Navigation Satellite System (GNSS). Researchers have used GNSS with real-time kinematic (RTK) system for high accuracy to achieve navigation for agricultural machinery \cite{thuilot2002automatic} and also robotic platforms \cite{bawden2017robot}. However GNSS has limitations of accuracy, possible outages, and reliance on geo-referenced auto-seeding. Hence the navigation requires perception information from local field structures \cite{ahmadi2020visual, billingsley1997successful}. 

Although RTK-GNSS based solutions had been popular in field robots, robust navigation in complex agricultural environments requires perception information from local field structures. In this direction, LiDAR was used for perception by Barawid \textit{et al.} \cite{barawid2007development} on orchards and by Malavazi \textit{et al.} \cite{malavazi2018lidar} in a simulated environment. Winterhalter \textit{et al.} \cite{winterhalter2018crop} used both LiDAR and RGB images to extract single lines in  row-crop fields with equal spacing.

Most of the earlier vision-based techniques use segmentation between plants and soil by applying some variation of greenness identification \textit{e.g.} excess green index (ExG) \cite{woebbecke1995color}. The lines required to extract paths from the segmented image have been estimated using techniques like Hough Transform \cite{marchant1995real, aastrand2005vision} or least squares fit \cite{zhang2018automated, garcia2018curved, bakken2019end}. However, such image processing based methods might fail in several situation \textit{e.g.}, plants covered with dirt after rainfall, ground covered with weeds or offshoots, different seasons, etc. Some other approaches involve using multi-spectral images \cite{haug2014plant} and plant stem emerging point (PSEP) using hand-crafted features \cite{midtiby2012estimating} for crop row location, however these methods cannot be generalized to mutliple crops. In a recent study, Ahmadi \textit{et al.} \cite{ahmadi2022towards} devised an image processing technique for crop center extraction using ExG followed by detecting individual crop-rows, achieving good performance on crops with both sparse and normal intensities. However, all the above techniques lack the ability to provide an overall scene understanding to facilitate multiple decision-making processes in a robot.

Supervised deep learning models for scene understanding have been successfully applied and popular in the autonomous navigation community as discussed in section \ref{sec:semseg-ad}. However there have been very limited work in agriculture. In a recent study, Song \textit{et al.} \cite{song2022navigation} used semantic segmentation, based on FCN, on wheat fields. However they only provide three classes (wheat, ground \& background), and do not provide fine pixel-wise annotations. Apart from wheat, it has been applied and tested for tea plantation \cite{lin2019development} and strawberry plantation \cite{ponnambalam2020autonomous}, which are relatively easier crops compared to wheat, corn, rice, canola, flax etc. In another study, Cao \textit{et al.} \cite{cao2022improved} proposed an improved ENet semantic segmentation network, followed by the random sampling consensus (RANSAC) algorithm to extract navigation lines. They evaluated the method on Crop Row Detection Lincoln Dataset (CRDLD) \cite{de2021towards} - a UAV dataset for sugar beet crop. However, the method does not provide pixel-wise labels for scene understanding, and the evaluations are limited to sparse and normal density crop rows. Bai \textit{et al.} \cite{bai2023vision} conducted a detailed and comprehensive review of vision-based navigation for agricultural autonomous vehicles and robots. DNN-based methods largely relies on annotated dataset but there are no good open-source benchmark datasets across multiple crops that can be used for development of semantic segmentation models for autonomous navigation in agriculture, the equivalent of Cityscapes dataset \cite{cordts2016cityscapes} on roads and streets. This motivates collection of our dataset, Agroscapes along with our transfer learning approach, in order to greatly reduce the amount of data required.

%-------------------------------------------------------------------------

\begin{table}
  \centering
  \begin{tabular}{@{}l|c|c|c@{}}
    \toprule
    Crop & Row-row & Crop & Camera   \\
    & [m] & Density & Placement \\
    \midrule
    Canola &  0.4 & normal, dense  & UGV, UAV, HH  \\ 
    Flax &  0.45 & normal, dense  & UGV, UAV, HH \\
    Strawberry & 0.3 & dense &  UAV, HH \\
    Bean &  0.5 & dense &  UAV, HH\\
    Corn & 0.4  & normal & UAV \\
    Cucumber & 0.5  & sparse & UAV \\
    \bottomrule
  \end{tabular}
  \caption{Distribution of data across crops with different row-to-row widths, crop density and camera placement. *HH = Handheld.}
  \label{tab:dataset}
\end{table}

\section{Dataset}
\label{sec:dataset}
\subsection{Data Collection}
\begin{figure}[t]
  \centering
    \begin{subfigure}{0.85\linewidth}
    \includegraphics[width=1\linewidth]{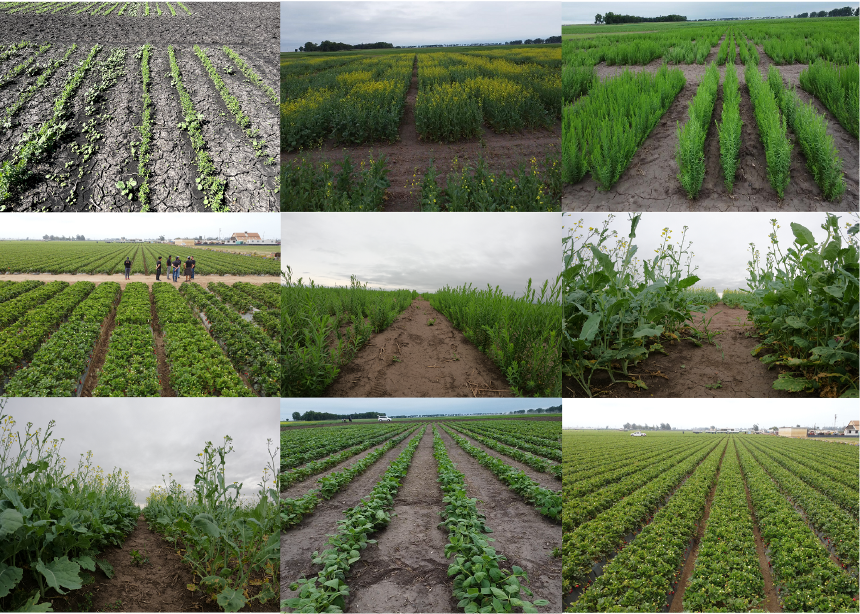}
    \captionsetup{belowskip=3pt, aboveskip=4pt}
    \caption{Agroscapes images from different crops and camera heights}
    \label{fig:agriscape-collage}
  \end{subfigure}
  \begin{subfigure}{0.85\linewidth}
    \includegraphics[width=1\linewidth]{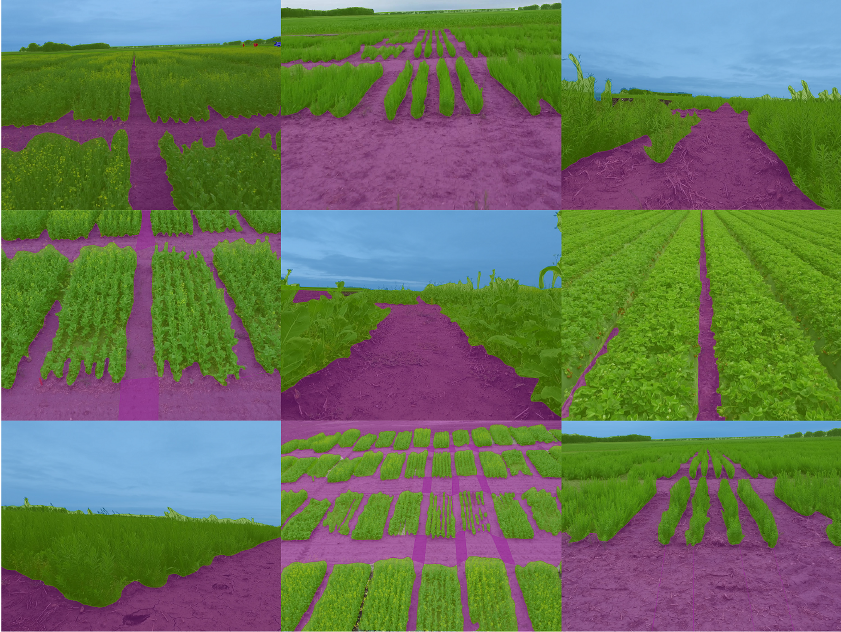}
    \captionsetup{belowskip=3pt, aboveskip=4pt}
    \caption{Fine pixel-wise annotations for scene understanding}
    \label{fig:semseg-collage}
  \end{subfigure}
   \begin{subfigure}{0.85\linewidth}
    \includegraphics[width=1\linewidth]{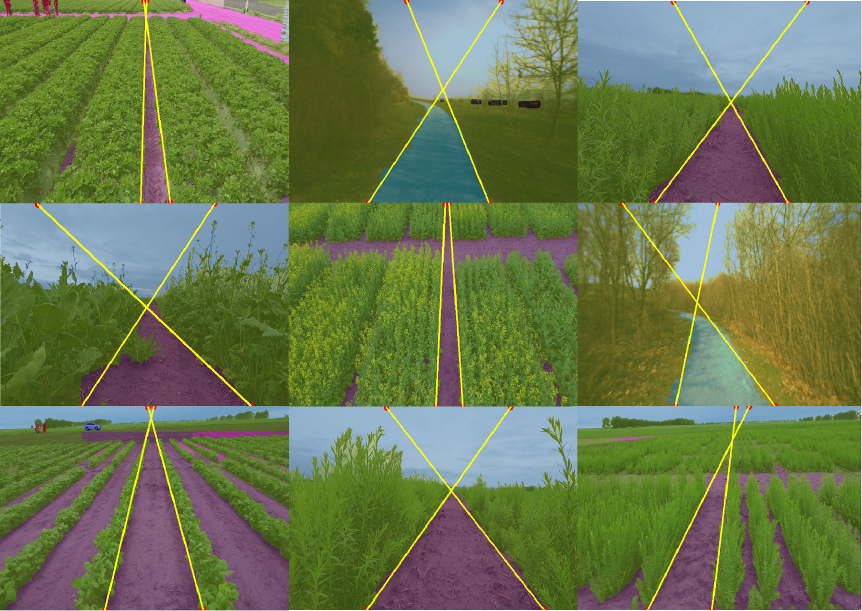}
    \captionsetup{belowskip=3pt, aboveskip=4pt}
    \caption{Line annotations for semantic line detection}
    \label{fig:line-anno-collage}
  \end{subfigure}
   \caption{An overview of Agroscapes dataset comprising different crops at different heights and subsequent annotations.}
   \label{fig:onecol}
\end{figure}

The data were collected in multiple locations across the United States for six different types of crops: strawberry, flax, canola, bean, corn and cucumber. Strawberry and cucumber data were collected in Oxnard, CA, while the data of the other crops were acquired in Fargo and Carrington, ND (see Figure \ref{fig:agriscape-collage}). The total data amounts to approximately 2,000 seconds of high-resolution video. Data were collected by cameras on three types of platforms: UAV, UGV and handheld (see Table \ref{tab:dataset}). In order to enhance the richness of the data, the viewpoint of the camera were varied in heights and angles for different iterations. This ensured that the collected data suited our initial purpose of developing an integrated framework applicable to different types of autonomous systems - mobile robots, drones and vehicles (e.g. autonomous tractors). The variations also provided our machine learning model more robustness for higher accuracy in varied environments.

% Explain the row-to-row distances in each crop
% Categorise into sparse, normal and intensive crops
% Or add a small table that summarises row-row distance, crop intensity, drone/ground images

\subsection{Pixel-wise Annotation}

Since one of the major goals of our work is to provide an open-source benchmark dataset in agriculture navigation for scene understanding, fine pixel-level annotations were carried out on the collected dataset (see Figure \ref{fig:semseg-collage}). The classes of the annotations were selected based on domain adaptation and the requirements of an agriculture scene. Hence a total of 120 images were finely annotated in 9 classes - soil, crop, weed, sky, human, vehicle, building, fence and other. Segmentation of soil and crop is the most important to determine the traversable and non-traversable regions. It is also important for the autonomous robot to identify and understand the presence of humans in its vicinity in an agricultural field. This is crucial for a safe-work environment with the workers on the field \cite{benos2020safety}. Other obstacles such as vehicles (or robots) and fences are also common to an agriculture environment. All images were annotated by a single annotator to ensure precise boundaries for each class and consistency among all annotations. CVAT (Computer Vision Annotation Tool) from Intel was used to execute all annotations. Among the annotated images, roughly 50\% are single-row images and the rest multiple-row images. Annotation of single-row images took approximately 20 minutes per image, whereas annotating multiple-row images took between 30 to 100 minutes per image.  Therefore, the total annotation effort for all images amounted to approximately 60 hours.

 % Benos \textit{et al.} \cite{benos2020safety} expresssed that due to automation and programmed robots, human safety is a concern and efforts must be taken to make robots extra sensitive to perceiving human proximity and ensure risk-free work environment.
\subsection{Line Annotation} 
The dataset used to train the semantic line detection model is a combination of our own data and the annotated Freiburg Forest dataset~\cite{valada2016}. To ensure that the model is trained on images similar to that of our test environment, we filter images from the Freiburg Forest dataset that are semantically similar to those taken from the agriculture field. The two main considerations for filtering were: 1) whether the boundary line that separates the crops from the traversable ground can be approximated using straight line, and 2) whether there were comparable amount of vegetation on the both sides of the two boundary lines. Once the RGB images have been selected, they were overlayed with their ground truth color masks, where the RGB images and the color masks have been weighed equally. Lastly, each weighted image was labeled with two semantic lines that mark the boundary between the crops and the ground (see Figure \ref{fig:line-anno-collage}). Each line was represented with the pixel coordinates of the endpoints that lie on the edges of the image. In total, 400 images were annotated and used for training. The ratio of ground images to aerial images was 1:1.

\section{Methodology}
\label{sec:methodology}

\begin{figure*}
  \centering
  \includegraphics[width=1\linewidth]{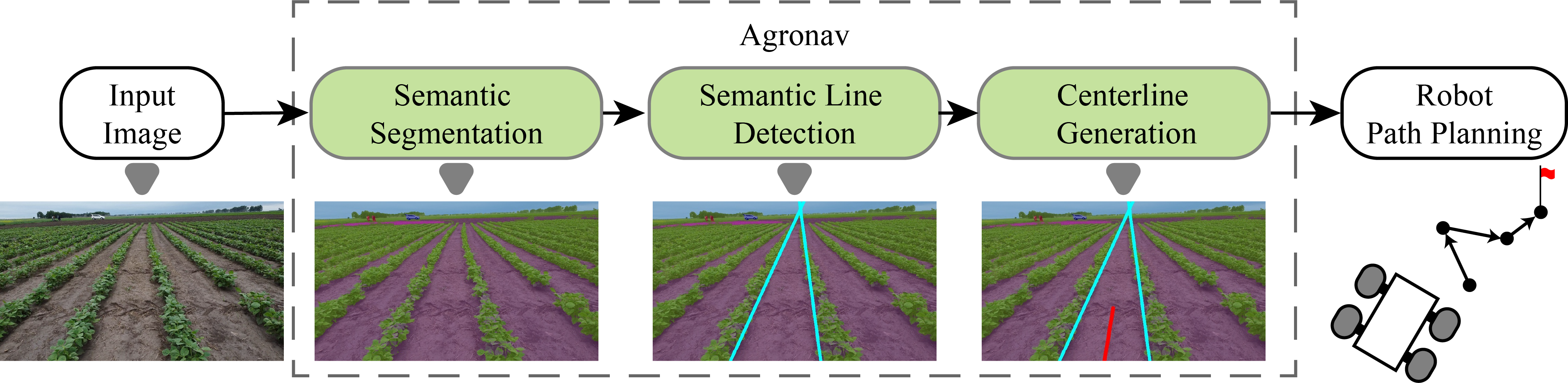}
  \caption{Overview of the Agronav pipeline with semantic segmentation, semantic line detection and centerline generation.}
  \label{fig:agronav-pipeline}
\end{figure*}

In this section we discuss the individual downstream tasks in the overall Agronav pipeline (see Figure \ref{fig:agronav-pipeline}).
\subsection{Semantic Segmentation}

\subsubsection{Models Selection}

Various semantic segmentation models were explored in the direction of transfer learning, using different pretrained checkpoints, to ensure good performance on Agroscapes dataset. Unsupervised domain adaptation inferences were executed on two kinds of models - large (high parameter) models  with limited real-time performance, and relatively smaller (low parameter) models, which were real-time capable. For the pretrained checkpoints, the models were experimented with checkpoints trained on ADE20K\cite{zhou2017scene}, Cityscapes\cite{cordts2016cityscapes}, COCO\cite{lin2014microsoft} and Pascal VOC\cite{everingham2009pascal}. The results showed that models pretrained on Cityscapes dataset achieved the best performance, which is reasonable, given the greater domain relevance of the Agroscapes dataset to the Cityscapes dataset compared to the other datasets. Among the large models, ViT-Adapter\cite{chen2022vision} was selected, which also has the state-of-the-art performance on the Cityscapes currently. Among the real-time models, three models - HRNet\cite{SunXLW19}, MobileNetV3\cite{howard2019searching} and ResNeSt\cite{zhang2020resnest} were selected for an ablation study on our domain adaptation. ViT-Adapter (ViT-A) had the best performance among the four models based on the inference study. 

% Mention the number of parameters in each model

HRNet\cite{SunXLW19} is a high-resolution network that is designed to preserve high-resolution representations throughout the network while maintaining a low computational cost. It employs a multi-resolution fusion approach that combines high-resolution and low-resolution representations to achieve both high accuracy and efficiency. MobileNet\cite{howard2019searching} is a lightweight network designed for mobile devices, which utilizes depthwise separable convolutions to reduce the number of parameters and computations. It has a small memory footprint and can achieve real-time performance on mobile devices. ResNeSt\cite{zhang2020resnest} is a recent advancement of the ResNet architecture that introduces nested and scale-specific feature aggregation to improve the model's ability to capture fine-grained patterns. It uses a split-attention mechanism to capture information from different feature maps and scales, resulting in improved accuracy on various computer vision tasks.

% Mention the architecture of ViT-adapter

\subsubsection{Supervised Domain Adaptation}

The classes of the pretrained checkpoints (on Cityscapes dataset) were reorganized for all the models. The Cityscapes checkpoints contain 19 classes, which were reorganized into 8 classes - soil, vegetation, sky, human, vehicle, building, fence and other. Now, the checkpoints were trained again on the source domain \textit{i.e.} Cityscapes based on these 8 labels. However, prior to retraining, the Cityscapes annotations had to be reorganized based on our domain relevance. Table \ref{tab:cityscapes-agroscapes} explains the reorganization from Cityscapes labels to the Agronav labels. This reorganization scheme guarantees inheritance of human recognition and obstacles avoidance capabilities of Cityscapes. 

\begin{table}
  \centering
  \begin{tabular}{@{}l|l@{}}
    \toprule
    Cityscapes & Agroscapes   \\
    \midrule
    Road, Sidewalk & Soil \\
    Vegetation, Terrain & Vegetation \\
    Sky & Sky \\
    Person, Rider & Human \\
    Building & Building \\
    Wall, Fence & Fence \\
    Car, Truck, Train, Bus, & Vehicle \\
    Motorcycle, Bicycle & \\
    Pole, Traffic Light, Traffic Signal  & Other \\
    % Traffic Signal & \\
    \bottomrule
  \end{tabular}
  \caption{Reorganization of Cityscapes labels for Agronav domain adaptation}
  \label{tab:cityscapes-agroscapes}
\end{table}

The final objective here is to achieve accurate, real-time semantic segmentation on the Agroscapes dataset using minimal number of annotated images (120 images). Considering the superior performance of the ViT-Adapter for zero-shot learning, given its large size, we first trained this large model on the annotated Agroscapes dataset. The resulting checkpoint was used to generate labels from 3850 unlabelled Agronav images. We visually inspected the generated labels and selected 1165 labels with mIoU of approximately 90\% or higher. These 1165 labels then served as the training data for the real-time models - HRNet, MobileNetV3 and ResNeSt. The models were trained in two stages: first, they were trained on the ViT-Adapter generated labels; then, fine-tuned by adding the manually annotated labels. In summary, our domain adaptation strategy leverages the high accuracy of the large model to improve the results of the real-time models, despite the limited availability of manually annotated data.

\subsection{Semantic Line Detection}

Our semantic line detection model is trained using the Deep Hough Transform method~\cite{zhao2021}. The pipeline includes four core components to detect semantic lines: 1) the feature pyramid network (FPN) extracts pixel-wise deep representations; 2) deep representations are converted from the spatial domain to the parameteric domain via Deep Hough Transform; 3) the line detector module to detect lines in the parametric space; 4) Reverse Hough Transform which converts the detected lines back into image space. In contrast to classical line detection algorithms which detect countless straight edges in an image, the Deep Hough Transform model can be explictly trained to output few most semantically meaningful lines. In addition, the lightweight of the trained semantic line detection model make it suitable for real-time applications. 

For autonomous crop-row navigation, every image consists of two most semantically meaningful lines. These lines mark the left-side and right-side boundaries of the traversable ground and the crops. The centerline, which serves as the ultimate guideline for navigation, can be extracted from these two lines. In an effort to develop a pipeline which is applicable for different scenarios, both ground and aerial robotic platforms, both single and multi-row images are annotated with two lines that mark the boundaries. As mentioned in Section \ref{sec:dataset}, line annotations are performed on the overlayed image, which is an equally weighted blend of the RGB image and the color mask. The latter is obtained as an output of the semantic segmentation model.

\subsection{Centerline Generation}

The final piece of our pipeline involves generating the centerline from the two output lines $l_1, l_2$ predicted by the semantic line detection model. Within the Deep Hough Transform framework, each line is parameterized by two parameters $r$ and $\theta$, where the distance parameter $r$ measures the distance between the line $l$ and the center of image, and the orientation parameter $\theta$ represents the angle between the line $l$ and the horizontal axis of the image. Following simple calculation, the entire set of pixels that correspond to each semantic line can be computed. We define the centerline as a set of midpoints of the two pixels $(x_{y_i}^{l_1}, y_{i})$ and $(x_{y_i}^{l_2}, y_{i})$, which reside on the same horizontal axis $y_i$. For simplicity, the centerlines were visualized for the bottom third of the image.
%-------------------------------------------------------------------------

\section{Results}
\label{sec:result}
\begin{figure*}
  \centering
  \includegraphics[width=1\linewidth]{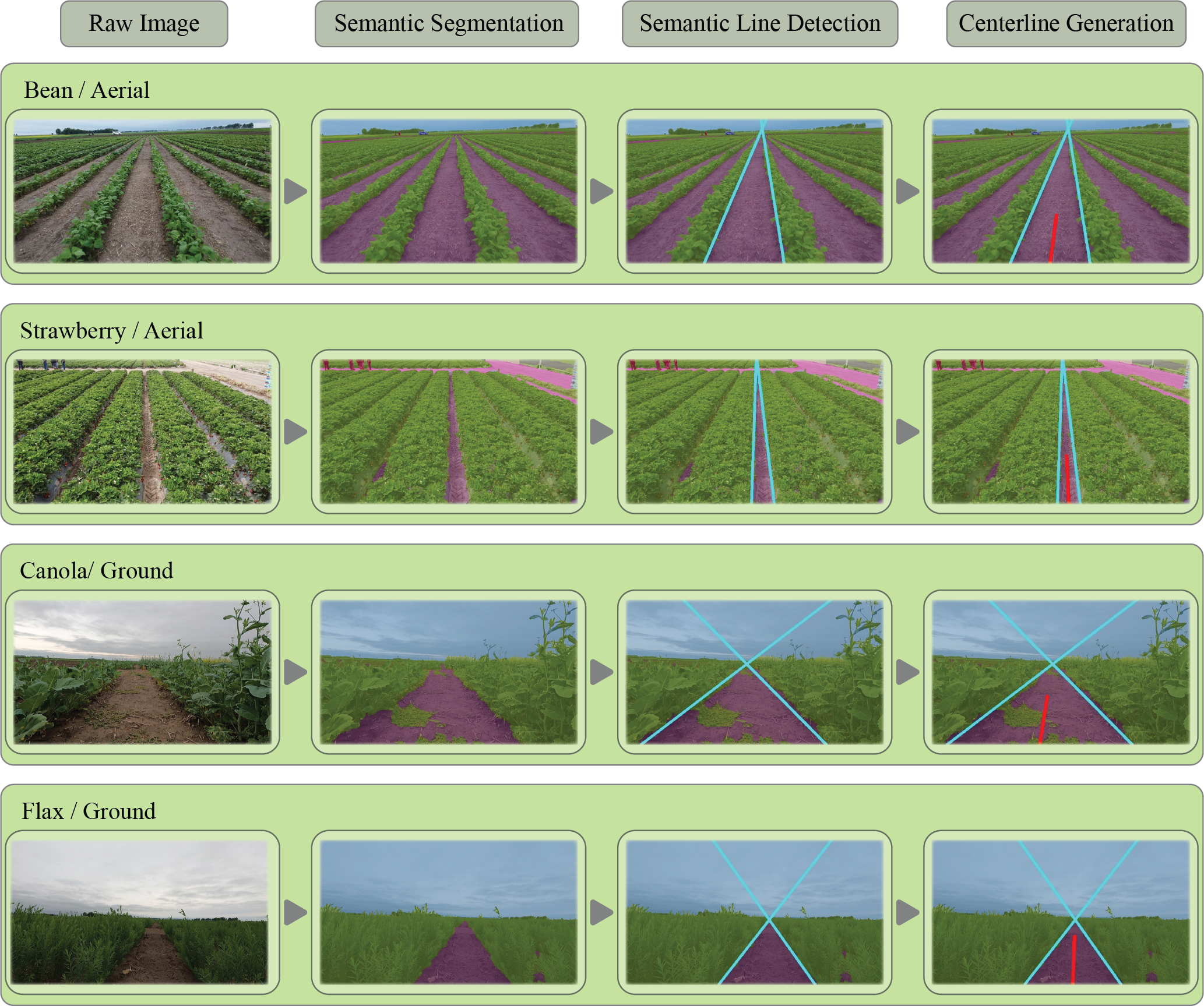}
  \caption{Demonstration of the end-to-end Agronav pipeline, tested on various crops.}
  \label{fig:testpipeline}
\end{figure*}

In this section we individually report the accuracy results from semantic segmentation and semantic line detection. We also report an ablation study of different models and techniques. Finally we provide visual assessment of the overall pipeline on unlabelled data (see Figure \ref{fig:testpipeline}).   
\subsection{Semantic Segmentation}

The training and validation procedures for all models were conducted on NVIDIA A100 GPUs. However, NVIDIA GeForce RTX 2080 GPUs were used for inference or testing on unlabelled images. This was done to evaluate the real-time performance of our models for deployment on mobile robots and other platforms. We used SGD as the optimizer for all models, with a learning rate of 0.01. Given the balanced distribution among the important classes - soil, vegetation and sky in most of the images, cross entropy loss was selected as our training loss. As for the evaluation metric, we used the most widely used mIoU (mean Intersection-over-Union) score.

% Mention the batch size and epochs

\begin{table}
  \centering
  \begin{tabular}{@{}l|c|ccc@{}}
    \toprule
    Method & mIoU & soil & vegetation & sky \\
    \midrule
    ViT-Adapter & \textbf{96.43 }& \textbf{94.24} & \textbf{96.72} & \textbf{98.23}\\
    HRNet & 95.28  & 92.39 & 95.48 & 97.92\\
    ResNeSt & 95.34 & 92.84 & 95.7 & 97.35 \\
    MobileNetV3 & 94.57 & 91.27 & 95.01 & 97.29\\
    \bottomrule
  \end{tabular}
  \captionsetup{belowskip=-5pt}
  \caption{Comparison of mIoU scores on Agroscapes dataset}
  \label{tab:ablation-study-1}
\end{table}

% As mentioned in sec. \ref{} all models were trained and validated with the manually labelled dataset. 
The mIoU scores of all models are reported in Table \ref{tab:ablation-study-1}. As the soil, vegetation, and sky classes represent the majority of the dataset and are the most important ones for our framework, we only report the mIoU scores for these classes. Our results show that the ViT-Adapter achieved the highest mIoU scores and is the most accurate model. The other real-time models are first trained on the ViT-A inferences and then fine-tuned on the manual labeled dataset. Among these models, ResNeSt achieved the best performance for soil, vegetation, and overall mIoU, while HRNet and MobileNetV3 also achieved equally good performance.

\begin{table}
  \centering
  \begin{tabular}{@{}l|c@{}}
    \toprule
    Method & FPS \\
    \midrule
    ViT-Adapter &  0.34 \\
    HRNet &  9.12 \\
    ResNeSt & 4.35 \\
    MobileNetV3 & \textbf{14.25} \\
    \bottomrule
  \end{tabular}
  \caption{Real time performance across all models}
  \label{tab:real-time}
\end{table}

We further evaluated the real-time performance of all models on NVIDIA GeForce RTX 2080 GPUs using video feeds from our dataset. The results in frames per second (FPS) are reported in Table \ref{tab:real-time}. Our results indicate that ViT-Adapter can not achieve a real-time performance, as it takes approximately 3 seconds to process a single frame. On the other hand, HRNet, ResNeSt, and MobileNetV3 all achieve good real-time performance, with MobileNetV3 achieving the highest FPS. However, the real-time performance for all models can be potentially improved by reducing the crop-size of the frames at the cost of accuracy.

\begin{table}
  \centering
  \begin{tabular}{@{}l|c|ccc@{}}
    \toprule
    Method & mIoU & soil & vegetation & sky \\
    \midrule
    \multicolumn{5}{l}{w/o ViT-A inferences}  \\
    \midrule
    HRNet &  92.62 & 92.86 & 89.43 & 96.64\\
    ResNeSt &  92.11 & 91.85 & 89.07 & 96.42\\
    MobileNetV3 & 92.48 & 92.45 & 89.35 & 96.7\\
    \midrule
    \multicolumn{5}{l}{with ViT-A inferences}  \\
    \midrule
    HRNet & 95.28  & 92.39 & 95.48 & \textbf{97.92}\\
    ResNeSt & \textbf{95.34} & \textbf{92.84} & \textbf{95.7} & 97.35 \\
    MobileNetV3 & 94.57 & 91.27 & 95.0 & 97.29\\
    \bottomrule
  \end{tabular}
  \captionsetup{belowskip=-5pt}
  \caption{Comparison of mIoU scores on the dataset with and without ViT-inferences}
  \label{tab:ViT-inferences}
\end{table}

To assess the impact of ViT-A inferences on model accuracy, we evaluated the three models with and without adding the inferences. The results are reported in Table \ref{tab:ViT-inferences}, which clearly show that the inferences significantly improved the mIoU scores for all three models. With ViT-A inferences we achieve superior performance, particularly an improved performance in segmenting vegetation. These findings demonstrate the success of our domain adaptation strategy. We also compare the performance of Agronav between ground and aerial images in Table \ref{tab:dataset-study}. It shows better performance with ground images primarily because it is more challenging to segment multiple soil boundaries from a height in aerial imagery. Nevertheless in a practical aerial operation, we can compromise some accuracy in segmenting soil and crop boundary.

\begin{table}
  \centering
  \begin{tabular}{@{}l|c|ccc@{}}
    \toprule
    Method & mIoU & soil & vegetation & sky \\
    \midrule
    \multicolumn{5}{l}{Ground}  \\
    \midrule
    HRNet &  96.54 & 95.77 & 96.31 & 97.63\\
    ResNeSt & 96.52 & \textbf{95.82} & 96.26 & 97.57\\
    MobileNetV3 & \textbf{96.66} & 95.74 & \textbf{96.43} & \textbf{97.88} \\
    \midrule
    \multicolumn{5}{l}{Aerial}  \\
    \midrule
    HRNet & \textbf{93.08} & 88.52 & \textbf{94.92} & \textbf{95.18} \\
    ResNeSt & 93.05 & \textbf{88.58} & 94.86 & 95.1 \\
    MobileNetV3 & 93.04 & 88.45 & 94.9 & 95.15 \\
    \bottomrule
  \end{tabular}
  \caption{Comparison of mIoU scores on ground vs. aerial images.}
  \label{tab:dataset-study}
\end{table}

\subsection{Semantic Line Detection}
%-------------------------------------------------------------------
We adopt the metrics proposed in ~\cite{zhao2021} to evaluate the similarity between a pair of predicted and ground truth lines, where the similarity between two lines is a function of the Euclidean distance and angular distance between the two lines. By representing the predicted lines and ground truth lines as each set of a Bipartite graph, the matching between the lines of two sets is done by solving the Maximum Bipartite Matching problem.

Following the matching process, true positives, false positives, and false negatives are identified, after which the Precision, Recall, and F-measure scores are evaluated. 
\begin{table}
  \centering
  \begin{tabular}{@{}l|c|c|c@{}}
    \toprule
    Method & Precision & Recall & F-measure \\
    \midrule
    Raw Images &  0.9021 & 0.7442 & 0.8156\\
    Overlayed Images & \textbf{0.9513} & \textbf{0.8984} & \textbf{0.9241}\\
    \bottomrule
  \end{tabular}
  \captionsetup{belowskip=-5pt}
  \caption{Comparison of the line detection model trained on raw and overlayed images.}
  \label{tab:overlayversusraw}
\end{table}

To assess the benefits of training the semantic line detection model on the overlayed images (an equally weighted blend of the color mask and raw RGB image), we compare it against a model trained on raw RGB images, as summarized in Table \ref{tab:overlayversusraw}. For fair comparison, all other training, optimizer, and dataset parameters were controlled. The semantic model trained on overlayed images trained more effectively, validating the reasoning behind the intended synergy between semantic segmentation and semantic line detection. 

\begin{table}
  \centering
  \begin{tabular}{@{}l|c|c|c@{}}
    \toprule
    Image type & Precision & Recall & F-measure \\
    \midrule
    Ground images &  0.9692 & 0.9692 & 0.9692\\
    Aerial Images &  0.7833 & 0.8616 & 0.8206\\
    \bottomrule
  \end{tabular}
  \caption{Performance evaluation of the line detection model on ground and aerial images.}
  \label{tab:groundversusdrone}
\end{table}

The results shown in Table \ref{tab:groundversusdrone} highlight good performance of the semantic line detection model on both ground and aerial images. The relatively inferior performance on aerial imagery can be attributed to the fact that aerial images include images with multiple crop rows, where multiple sets of boundary lines can be drawn, hence making the task more difficult.

\section{Conclusion}
\label{sec:conclusion}

In this study, an end-to-end framework for vision-based autonomous navigation in agricultural fields was proposed. By adopting domain adaptation for semantic segmentation, we have trained a robust segmentation model that can successfully segment an image into 8 classes including soil, vegetation, sky, human, vehicle, building, fence, and other. The original 19 classes of the Cityscapes dataset were reorganized and then merged with our own data, collected across six different crops. Additionally, we have demonstrated the effectiveness of the semantic line detection model for detecting boundary lines, which capitalizes on the inherent structure of the field where crops are planted in straight rows. Furthermore, the study validates the synergy between semantic segmentation and semantic line detection by showing that the semantic line detection model trains better on overlayed images. We are also releasing Agroscapes - an open-source dataset for scene understanding in agriculture. This dataset is expected to serve as a valuable resource for studying autonomous navigation in agricultural fields and as a benchmark for future research. 

In future work, the results of this study will be applied to a physical robotic platform with dimensions of 75 in (L) x 63 in (W) x 40 in (H), which was initially designed for weed management and data collection in flax and canola fields. The use of this platform will allow for the collection of field trial data. In addition, we hope to release more labeled and unlabeled images in our open-source dataset, Agroscapes, collected across more crops. We also aim to incorporate segmentation of weeds from the crops in future semantic segmentation models.
%-------------------------------------------------------------------------
% \section{References}

% \begin{table}
%   \centering
%   \begin{tabular}{@{}lc@{}}
%     \toprule
%     Method & Frobnability \\
%     \midrule
%     Theirs & Frumpy \\
%     Yours & Frobbly \\
%     Ours & Makes one's heart Frob\\
%     \bottomrule
%   \end{tabular}
%   \caption{Results.   Ours is better.}
%   \label{tab:example}
% \end{table}

%-------------------------------------------------------------------------
\section*{Acknowledgement}
We acknowledge the funding and support from United States Department of Agriculture (USDA Award No. 2021-67022-34200 \& 2022-67022-37021) and National Science Foundation (NSF Award No. IIS-2047663 \& CNS-2213839).
%%%%%%%%% REFERENCES


{\small
\begin{thebibliography}{10}\itemsep=-1pt

\bibitem{aghi2020}
Diego Aghi, Vittorio Mazzia, and Marcello Chiaberge.
\newblock Local motion planner for autonomous navigation in vineyards with a
  rgb-d camera-based algorithm and deep learning synergy.
\newblock {\em Machines}, 8(2), 2020.

\bibitem{ahmadi2022towards}
Alireza Ahmadi, Michael Halstead, and Chris McCool.
\newblock Towards autonomous visual navigation in arable fields.
\newblock In {\em 2022 IEEE/RSJ International Conference on Intelligent Robots
  and Systems (IROS)}, pages 6585--6592. IEEE, 2022.

\bibitem{ahmadi2020visual}
Alireza Ahmadi, Lorenzo Nardi, Nived Chebrolu, and Cyrill Stachniss.
\newblock Visual servoing-based navigation for monitoring row-crop fields.
\newblock In {\em 2020 IEEE International Conference on Robotics and Automation
  (ICRA)}, pages 4920--4926. IEEE, 2020.

\bibitem{aastrand2005vision}
Bj{\"o}rn {\AA}strand and Albert-Jan Baerveldt.
\newblock A vision based row-following system for agricultural field machinery.
\newblock {\em Mechatronics}, 15(2):251--269, 2005.

\bibitem{bai2023vision}
Yuhao Bai, Baohua Zhang, Naimin Xu, Jun Zhou, Jiayou Shi, and Zhihua Diao.
\newblock Vision-based navigation and guidance for agricultural autonomous
  vehicles and robots: A review.
\newblock {\em Computers and Electronics in Agriculture}, 205:107584, 2023.

\bibitem{bakken2019end}
Marianne Bakken, Richard~JD Moore, and P{\aa}l From.
\newblock End-to-end learning for autonomous crop row-following.
\newblock {\em IFAC-PapersOnLine}, 52(30):102--107, 2019.

\bibitem{barawid2007development}
Oscar~C Barawid~Jr, Akira Mizushima, Kazunobu Ishii, and Noboru Noguchi.
\newblock Development of an autonomous navigation system using a
  two-dimensional laser scanner in an orchard application.
\newblock {\em Biosystems Engineering}, 96(2):139--149, 2007.

\bibitem{bawden2017robot}
Owen Bawden, Jason Kulk, Ray Russell, Chris McCool, Andrew English, Feras
  Dayoub, Chris Lehnert, and Tristan Perez.
\newblock Robot for weed species plant-specific management.
\newblock {\em Journal of Field Robotics}, 34(6):1179--1199, 2017.

\bibitem{benos2020safety}
Lefteris Benos, Avital Bechar, and Dionysis Bochtis.
\newblock Safety and ergonomics in human-robot interactive agricultural
  operations.
\newblock {\em Biosystems Engineering}, 200:55--72, 2020.

\bibitem{bjornlund2022}
Vibeke Bjornlund, Henning Bjornlund, and Andr{\'e} van Rooyen.
\newblock Why food insecurity persists in sub-saharan africa: A review of
  existing evidence.
\newblock {\em Food Security}, 14(4):845--864, 2022.

\bibitem{brown2019}
Thomas~C Brown, Vinod Mahat, and Jorge~A Ramirez.
\newblock Adaptation to future water shortages in the united states caused by
  population growth and climate change.
\newblock {\em Earth's Future}, 7(3):219--234, 2019.

\bibitem{cao2022improved}
Maoyong Cao, Fangfang Tang, Peng Ji, and Fengying Ma.
\newblock Improved real-time semantic segmentation network model for crop
  vision navigation line detection.
\newblock {\em Frontiers in Plant Science}, 13, 2022.

\bibitem{chen2017deeplab}
Liang-Chieh Chen, George Papandreou, Iasonas Kokkinos, Kevin Murphy, and Alan~L
  Yuille.
\newblock Deeplab: Semantic image segmentation with deep convolutional nets,
  atrous convolution, and fully connected crfs.
\newblock {\em IEEE transactions on pattern analysis and machine intelligence},
  40(4):834--848, 2017.

\bibitem{chen2022vision}
Zhe Chen, Yuchen Duan, Wenhai Wang, Junjun He, Tong Lu, Jifeng Dai, and Yu
  Qiao.
\newblock Vision transformer adapter for dense predictions.
\newblock {\em arXiv preprint arXiv:2205.08534}, 2022.

\bibitem{CISTERNAS2020}
Isabel Cisternas, Ignacio VelÃ¡squez, AngÃ©lica Caro, and Alfonso RodrÃ­guez.
\newblock Systematic literature review of implementations of precision
  agriculture.
\newblock {\em Computers and Electronics in Agriculture}, 176:105626, 2020.

\bibitem{cordts2016cityscapes}
Marius Cordts, Mohamed Omran, Sebastian Ramos, Timo Rehfeld, Markus Enzweiler,
  Rodrigo Benenson, Uwe Franke, Stefan Roth, and Bernt Schiele.
\newblock The cityscapes dataset for semantic urban scene understanding.
\newblock In {\em Proceedings of the IEEE conference on computer vision and
  pattern recognition}, pages 3213--3223, 2016.

\bibitem{de2021towards}
Rajitha de Silva, Grzegorz Cielniak, and Junfeng Gao.
\newblock Towards agricultural autonomy: crop row detection under varying field
  conditions using deep learning.
\newblock {\em arXiv preprint arXiv:2109.08247}, 2021.

\bibitem{dosovitskiy2020image}
Alexey Dosovitskiy, Lucas Beyer, Alexander Kolesnikov, Dirk Weissenborn,
  Xiaohua Zhai, Thomas Unterthiner, Mostafa Dehghani, Matthias Minderer, Georg
  Heigold, Sylvain Gelly, et~al.
\newblock An image is worth 16x16 words: Transformers for image recognition at
  scale.
\newblock {\em arXiv preprint arXiv:2010.11929}, 2020.

\bibitem{emmi2021}
L Emmi, E Le~Fl{\'e}cher, V Cadenat, and M Devy.
\newblock A hybrid representation of the environment to improve autonomous
  navigation of mobile robots in agriculture.
\newblock {\em Precision Agriculture}, 22:524--549, 2021.

\bibitem{everingham2009pascal}
Mark Everingham, Luc Van~Gool, Christopher~KI Williams, John Winn, and Andrew
  Zisserman.
\newblock The pascal visual object classes (voc) challenge.
\newblock {\em International journal of computer vision}, 88:303--308, 2009.

\bibitem{garcia2018curved}
Iv{\'a}n Garc{\'\i}a-Santill{\'a}n, Jos{\'e}~Miguel Guerrero, Mart{\'\i}n
  Montalvo, and Gonzalo Pajares.
\newblock Curved and straight crop row detection by accumulation of green
  pixels from images in maize fields.
\newblock {\em Precision Agriculture}, 19(1):18--41, 2018.

\bibitem{geiger2012we}
Andreas Geiger, Philip Lenz, and Raquel Urtasun.
\newblock Are we ready for autonomous driving? the kitti vision benchmark
  suite.
\newblock In {\em 2012 IEEE conference on computer vision and pattern
  recognition}, pages 3354--3361. IEEE, 2012.

\bibitem{haug2014plant}
Sebastian Haug, Peter Biber, Andreas Michaels, and J{\"o}rn Ostermann.
\newblock Plant stem detection and position estimation using machine vision.
\newblock In {\em Workshop Proc. of Conf. on Intelligent Autonomous Systems
  (IAS)}, pages 483--490, 2014.

\bibitem{howard2019searching}
Andrew Howard, Mark Sandler, Grace Chu, Liang-Chieh Chen, Bo Chen, Mingxing
  Tan, Weijun Wang, Yukun Zhu, Ruoming Pang, Vijay Vasudevan, et~al.
\newblock Searching for mobilenetv3.
\newblock In {\em Proceedings of the IEEE/CVF international conference on
  computer vision}, pages 1314--1324, 2019.

\bibitem{iqbal2020}
Jawad Iqbal, Rui Xu, Shangpeng Sun, and Changying Li.
\newblock Simulation of an autonomous mobile robot for lidar-based in-field
  phenotyping and navigation.
\newblock {\em Robotics}, 9(2), 2020.

\bibitem{jin2020}
Dongkwon Jin, Jun-Tae Lee, and Chang-Su Kim.
\newblock Semantic line detection using mirror attention and comparative
  ranking and matching.
\newblock In {\em Computer Vision--ECCV 2020: 16th European Conference,
  Glasgow, UK, August 23--28, 2020, Proceedings, Part XX 16}, pages 119--135.
  Springer, 2020.

\bibitem{laborde2020}
David Laborde, Will Martin, Johan Swinnen, and Rob Vos.
\newblock Covid-19 risks to global food security.
\newblock {\em Science}, 369(6503):500--502, 2020.

\bibitem{Lee2017}
Jun-Tae Lee, Han-Ul Kim, Chul Lee, and Chang-Su Kim.
\newblock Semantic line detection and its applications.
\newblock In {\em 2017 IEEE International Conference on Computer Vision
  (ICCV)}, pages 3249--3257, 2017.

\bibitem{lin2014microsoft}
Tsung-Yi Lin, Michael Maire, Serge Belongie, James Hays, Pietro Perona, Deva
  Ramanan, Piotr Doll{\'a}r, and C~Lawrence Zitnick.
\newblock Microsoft coco: Common objects in context.
\newblock In {\em Computer Vision--ECCV 2014: 13th European Conference, Zurich,
  Switzerland, September 6-12, 2014, Proceedings, Part V 13}, pages 740--755.
  Springer, 2014.

\bibitem{lin2019development}
Yu-Kai Lin and Shih-Fang Chen.
\newblock Development of navigation system for tea field machine using semantic
  segmentation.
\newblock {\em IFAC-PapersOnLine}, 52(30):108--113, 2019.

\bibitem{liu2021Swin}
Ze Liu, Yutong Lin, Yue Cao, Han Hu, Yixuan Wei, Zheng Zhang, Stephen Lin, and
  Baining Guo.
\newblock Swin transformer: Hierarchical vision transformer using shifted
  windows.
\newblock In {\em Proceedings of the IEEE/CVF international conference on
  computer vision}, pages 10012--10022, 2021.

\bibitem{lohar2021}
Shreya Lohar, Lei Zhu, Stanley Young, Peter Graf, and Michael Blanton.
\newblock Sensing technology survey for obstacle detection in vegetation.
\newblock {\em Future Transportation}, 1(3):672--685, 2021.

\bibitem{long2015fully}
Jonathan Long, Evan Shelhamer, and Trevor Darrell.
\newblock Fully convolutional networks for semantic segmentation.
\newblock In {\em Proceedings of the IEEE conference on computer vision and
  pattern recognition}, pages 3431--3440, 2015.

\bibitem{malavazi2018lidar}
Flavio~BP Malavazi, Remy Guyonneau, Jean-Baptiste Fasquel, Sebastien Lagrange,
  and Franck Mercier.
\newblock Lidar-only based navigation algorithm for an autonomous agricultural
  robot.
\newblock {\em Computers and electronics in agriculture}, 154:71--79, 2018.

\bibitem{marchant1995real}
John~A Marchant and Renaud Brivot.
\newblock Real-time tracking of plant rows using a hough transform.
\newblock {\em Real-time imaging}, 1(5):363--371, 1995.

\bibitem{midtiby2012estimating}
Henrik~S Midtiby, Thomas~M Giselsson, and Rasmus~N J{\o}rgensen.
\newblock Estimating the plant stem emerging points (pseps) of sugar beets at
  early growth stages.
\newblock {\em Biosystems engineering}, 111(1):83--90, 2012.

\bibitem{monteiro2011}
AntÃ³nio Monteiro, SÃ©rgio Santos, and Pedro GonÃ§alves.
\newblock Precision agriculture for crop and livestock farmingâ€”brief review.
\newblock {\em Animals}, 11(8), 2021.

\bibitem{ponnambalam2020autonomous}
Vignesh~Raja Ponnambalam, Marianne Bakken, Richard~JD Moore, Jon Glenn
  Omholt~Gjevestad, and P{\aa}l Johan~From.
\newblock Autonomous crop row guidance using adaptive multi-roi in strawberry
  fields.
\newblock {\em Sensors}, 20(18):5249, 2020.

\bibitem{ronneberger2015u}
Olaf Ronneberger, Philipp Fischer, and Thomas Brox.
\newblock U-net: Convolutional networks for biomedical image segmentation.
\newblock In {\em Medical Image Computing and Computer-Assisted
  Intervention--MICCAI 2015: 18th International Conference, Munich, Germany,
  October 5-9, 2015, Proceedings, Part III 18}, pages 234--241. Springer, 2015.

\bibitem{shafi2019}
Uferah Shafi, Rafia Mumtaz, Jos{\'e} Garc{\'\i}a-Nieto, Syed~Ali Hassan, Syed
  Ali~Raza Zaidi, and Naveed Iqbal.
\newblock Precision agriculture techniques and practices: From considerations
  to applications.
\newblock {\em Sensors}, 19(17):3796, 2019.

\bibitem{song2022navigation}
Yan Song, Feiyang Xu, Qi Yao, Jialin Liu, and Shuai Yang.
\newblock Navigation algorithm based on semantic segmentation in wheat fields
  using an rgb-d camera.
\newblock {\em Information Processing in Agriculture}, 2022.

\bibitem{SunXLW19}
Ke Sun, Bin Xiao, Dong Liu, and Jingdong Wang.
\newblock Deep high-resolution representation learning for human pose
  estimation.
\newblock In {\em CVPR}, 2019.

\bibitem{thuilot2002automatic}
Beno{\^\i}t Thuilot, Christophe Cariou, Philippe Martinet, and Michel Berducat.
\newblock Automatic guidance of a farm tractor relying on a single cp-dgps.
\newblock {\em Autonomous robots}, 13(1):53--71, 2002.

\bibitem{valada2016}
Abhinav Valada, Gabriel Oliveira, Thomas Brox, and Wolfram Burgard.
\newblock Deep multispectral semantic scene understanding of forested
  environments using multimodal fusion.
\newblock In {\em International Symposium on Experimental Robotics (ISER)},
  2016.

\bibitem{verschuur2021}
Jasper Verschuur, Sihan Li, Piotr Wolski, and Friederike~EL Otto.
\newblock Climate change as a driver of food insecurity in the 2007
  lesotho-south africa drought.
\newblock {\em Scientific reports}, 11(1):3852, 2021.

\bibitem{winterhalter2018crop}
Wera Winterhalter, Freya~Veronika Fleckenstein, Christian Dornhege, and Wolfram
  Burgard.
\newblock Crop row detection on tiny plants with the pattern hough transform.
\newblock {\em IEEE Robotics and Automation Letters}, 3(4):3394--3401, 2018.

\bibitem{woebbecke1995color}
David~M Woebbecke, George~E Meyer, Kenneth Von~Bargen, and David~A Mortensen.
\newblock Color indices for weed identification under various soil, residue,
  and lighting conditions.
\newblock {\em Transactions of the ASAE}, 38(1):259--269, 1995.

\bibitem{xie2021segformer}
Enze Xie, Wenhai Wang, Zhiding Yu, Anima Anandkumar, Jose~M Alvarez, and Ping
  Luo.
\newblock Segformer: Simple and efficient design for semantic segmentation with
  transformers.
\newblock {\em Advances in Neural Information Processing Systems},
  34:12077--12090, 2021.

\bibitem{zhang2020resnest}
Hang Zhang, Chongruo Wu, Zhongyue Zhang, Yi Zhu, Zhi Zhang, Haibin Lin, Yue
  Sun, Tong He, Jonas Muller, R. Manmatha, Mu Li, and Alexander Smola.
\newblock Resnest: Split-attention networks.
\newblock {\em arXiv preprint arXiv:2004.08955}, 2020.

\bibitem{zhang2018automated}
Xiya Zhang, Xiaona Li, Baohua Zhang, Jun Zhou, Guangzhao Tian, Yingjun Xiong,
  and Baoxing Gu.
\newblock Automated robust crop-row detection in maize fields based on position
  clustering algorithm and shortest path method.
\newblock {\em Computers and electronics in agriculture}, 154:165--175, 2018.

\bibitem{zhao2021}
Kai Zhao, Qi Han, Chang-Bin Zhang, Jun Xu, and Ming-Ming Cheng.
\newblock Deep hough transform for semantic line detection.
\newblock {\em IEEE Transactions on Pattern Analysis and Machine Intelligence},
  44(9):4793--4806, 2021.

\bibitem{zhou2017scene}
Bolei Zhou, Hang Zhao, Xavier Puig, Sanja Fidler, Adela Barriuso, and Antonio
  Torralba.
\newblock Scene parsing through ade20k dataset.
\newblock In {\em Proceedings of the IEEE conference on computer vision and
  pattern recognition}, pages 633--641, 2017.

\end{thebibliography}

}
\end{document}